\documentclass[
]{ceurart}

\usepackage{graphicx}
\usepackage{svg}
\usepackage{amsmath}
\usepackage{caption}
\usepackage{subcaption}
\usepackage{wrapfig}
\usepackage{todonotes}
\usepackage{url}

\begin{document}

\copyrightyear{2024}
\copyrightclause{Copyright for this paper by its authors.
  Use permitted under Creative Commons License Attribution 4.0
  International (CC BY 4.0).}

\conference{WISDOMS 2024 – First International Workshop on Integrating the Semantics of Data, Ontologies, Moral and cultural values and their Societal impact. Co-located with ESWC 2024}

\title{Explainable Moral Values: a neuro-symbolic approach to value classification}

\author[1]{Nicolas Lazzari}[%
orcid=0000-0002-1601-7689,
email=nicolas.lazzar3@unibo.it,
]
\cormark[1]

\author[2]{Stefano {De Giorgis}}[%
orcid=0000-0003-4133-3445,
email=stefano.degiorgis2@unibo.it,
]
\cormark[1]

\author[2, 3]{Aldo Gangemi}[%
orcid=0000-0001-5568-2684,
email=aldo.gangemi@unibo.it
]

\author[4]{Valentina Presutti}[%
orcid=0000-0002-9380-5160,
email=valentina.presutti@unibo.it
]

\address[1]{Department of Computer Science, \textit{University of Pisa}, Italy}
\address[2]{Institute of Cognitive Sciences and Technologies, \textit{National Research Council}, Italy}
\address[3]{Department of Philosophy, \textit{University of Bologna}, Bologna, Italy}
\address[4]{Department of Modern Languages, Literature, and Cultures, \textit{University of Bologna}, Italy}
\cortext[1]{Corresponding authors.}

\begin{abstract}
This work explores the integration of ontology-based reasoning and Machine Learning techniques for explainable value classification. By relying on an ontological formalization of moral values as in the Moral Foundations Theory, relying on the DnS Ontology Design Pattern, the \textit{sandra} neuro-symbolic reasoner is used to infer values (fomalized as descriptions) that are \emph{satisfied by} a certain sentence. Sentences, alongside their structured representation, are automatically generated using an open-source Large Language Model. The inferred descriptions are used to automatically detect the value associated with a sentence. We show that only relying on the reasoner's inference results in explainable classification comparable to other more complex approaches. We show that combining the reasoner's inferences with distributional semantics methods largely outperforms all the baselines, including complex models based on neural network architectures. Finally, we build a visualization tool to explore the potential of theory-based values classification, which is publicly available at \url{http://xmv.geomeaning.com/}.
\end{abstract}


\maketitle
    
    \section{Introduction}
    \label{sec:intro}

The recent advent of AI models showing impressive performance in generative tasks, especially related to linguistic and visual production, have made the Value Alignment \cite{sierra2021value,fisac2020pragmatic} problem even more relevant for the close future.
Detecting and classifying human values in multimodal data has emerged as a crucial task in various domains, especially with the proven capabilities by Large Language Models (LLMs) where aligning them with human beliefs is gaining mainstream attention \cite{ji2023llmalign}. The ability to automatically discern values from text holds significance for understanding human behavior, decision-making processes, and societal trends . However, achieving this classification presents several challenges.

Firstly, value classification is inherently subjective and context-dependent, making it difficult to objectively identify a unique value associated with a given sentence. Additionally, it's imperative to consider diverse perspectives when undertaking this task to ensure a comprehensive understanding of the underlying values.

Current approaches predominantly rely on Machine Learning (ML) techniques, particularly pre-trained LLMs, for value classification. While these methods demonstrate effectiveness, the opacity of their decision-making processes raises concerns regarding the semantic understanding of sentences versus mere pattern matching. As such, there is a pressing need for an approach that not only provides accurate predictions but also offers interpretability, allowing stakeholders to assess the alignment of predictions with established knowledge of values.

To address these challenges, we propose a novel framework that integrates ontology-based reasoning with ML techniques. Specifically, we leverage the \emph{sandra} neuro-symbolic reasoner, which utilizes the DnS Ontology Design Pattern, to infer the various perspectives from which a given situation, represented as a sentence, can be interpreted. In our framework, we focus on Moral Foundations Theory by Graham and Haidt \cite{graham2013moral} as the theoretical framework guiding our ontology formalization.

A critical aspect of our approach is the structured representation of sentences, which we obtain using an open-source LLM. Through empirical evaluation, we demonstrate the feasibility and efficacy of this approach, showing desirable properties such as the correlation between detected emotion polarity in sentences and the polarity of associated values.

We establish a robust baseline for value classification by solely relying on the reasoner's inferences, demonstrating comparable performance to more complex approaches involving pre-trained LLMs. Importantly, our method ensures explainability, enabling us to provide justifications for classification decisions.

Furthermore, we showcase the synergistic benefits of combining our approach with other NLP techniques, resulting in significant performance improvements over all baselines, including LLM-based methods, without sacrificing interpretability. Through this introduction, we set the stage for the subsequent sections, where we delve deeper into the methodology, experiments, and results of our proposed framework for explainable value classification.

The paper is organized as follows: Section \ref{sec:related} presents the related works in the human value fields. Section \ref{sec:data} describes the approach used to produce the dataset. Section \ref{sec:background} provides an overview of the neuro-symbolic techniques used to perform inferences over the dataset while Section \ref{sec:method} presents the experiments performed and discuss its results, showcasing the interpretability of some methods. Finally, Section \ref{sec:conclusion} provides an overview of the contribution and identify limitations that will be explored in future works.

    \section{Related Works}
    \label{sec:related}
In this section, we review existing literature relevant to the integration of ontology-based reasoning and Machine Learning (ML) techniques for value formalization and classification, focusing on the Moral Foundations Theory, detailing our formalization via reusing the DnS Ontology Design Pattern.

\paragraph{Values Theoretical Frameworks}
Value formalization, intended as the construction of a theoretical structure encompassing a set of entities, and consequent elicitation of semantic relation intertwining moral and cultural values has a long and cross disciplinary background. Here we focus on the Moral Foundations Theory (MFT), as detailed in the following, but other existing theoretical frameworks are Basic Human Values, by Scwhartz \cite{schwartz_extending_2001,gimenez_analysis_2019}, and Morality as Cooperation \cite{curry2016morality,curry2022moral}.

From an knowledge representation perspective there have been a previous attempt to represent the notion of value: the Ethics Ontology \cite{debellis15327812ethics}, which explore the commonsense notion of ``value'' to a certain extent, but lacks the grounding to a specific theory. Another relevant work is the Value Awareness Ontology \cite{holgado2024ontology}, focused on agent-based value-aware and normative systems. In this paper we extended previous existing work, focused on the above mentioned theories as ontological modules: the ValueNet ontology \cite{degiorgis2022valuenet}.
ValueNet is an ontology network formalising several theories, and modeling moral and cultural values as semantic frames \cite{fillmore1982framsemantics,minsky1974framework}.
The frame-based modeling commits to two main assumptions: (i) Values are modeled as n-ary relations, having roles playing a specific semantic function (e.g. Agent, Undergoer, Victim, Beneficiary, etc.) and (ii) values can be ``evoked'' by a set of lexical units, which could not be understood without the whole frame structure (e.g. the notion of ``Victim'' makes sense only if there is some form of Harm perpetrated to the victim, and therefore the lexical unit ``victim'', finds its meaning in this implicit semantic load.


\paragraph{Values Detection and Classification}
In the realm of value classification, current approaches predominantly rely on ML techniques, with BERT-based models often considered state-of-the-art \cite{mirzakhmedova2023touch}. 
However, these methods typically lack transparency in their decision-making process, posing challenges for explainability. Furthermore, datasets for training such models are often constructed through manual annotation or extraction from existing sources, which may introduce biases. 
Previous works like \cite{asprino2022uncovering,bulla2022detection} focus on more explainable methodologies, reusing semantic frames activation to detect values from MFT.

Notably, our approach differs by utilizing a generative method to automatically create structured representations of sentences. Importantly, our approach serves as a benchmark for evaluating the quality of classification methods, given the absence of human evaluation.
While attempts at explainable value classification are emerging, they often lack a robust theoretical foundation. Our approach combines ML techniques with ontology-based reasoning, leveraging Moral Foundations Theory formalization, to provide both performance and interpretability. This integration allows us to anchor our classification framework in established and well-understood theories, setting our approach apart from previous methods.

\paragraph{Moral Foundations Theory}
MFT is a proposal for a theory of social, cultural and moral behaviours. It aims at being universal and not depending on culture - at least in its dyadic oppositional structure of value vs violation. It considers values, intended as ``foundations'', as universal, while their real world occurrence is cultural and context-specific, namely e.g. the concept of ``Care'' is pan-cultural, while its actual real world realization (or its violation) is cultural dependent.
MFT organises values and violations in terms of dyadic oppositions, and it explains each dyad in terms of behavioural cognitivism, namely, e.g. \textit{Care vs Harm} is grounded in the attachment systems and some form of empathy, intended as the ability to not only understand, but also feel the same feelings as others, being able to imagine hypothetical scenarios, in which we are living some positive or negative mental or physical state, which we actually don't live.
The other dyads are Fairness/Cheating, Loyalty/Betrayal, Authority/Subversion, Sanctity/Degradation, Liberty/Oppression.
Original ValueNet MFT ontological module includes one class per each of these value-violations, and aligns it, following frame semantics principles, to entities from FrameNet \cite{baker_berkeley_nodate}, WordNet \cite{miller_wordnet_1995}, VerbNet \cite{schuler2005verbnet}, PropBank \cite{kingsbury_treebank_nodate} and several others in the Framester hub \cite{gangemi2016framester}.
However, in this work, we use only those value-frames which are evoked by some existing frame in the Framester hub, and for which we can reuse the semantic roles, resulting in a final set of 10 value-frames out of the original 12.


    
    \section{Synthetic Data Generation}
    \label{sec:data}
\begin{figure}[htbp]
    \centering
    \includegraphics[width=\textwidth]{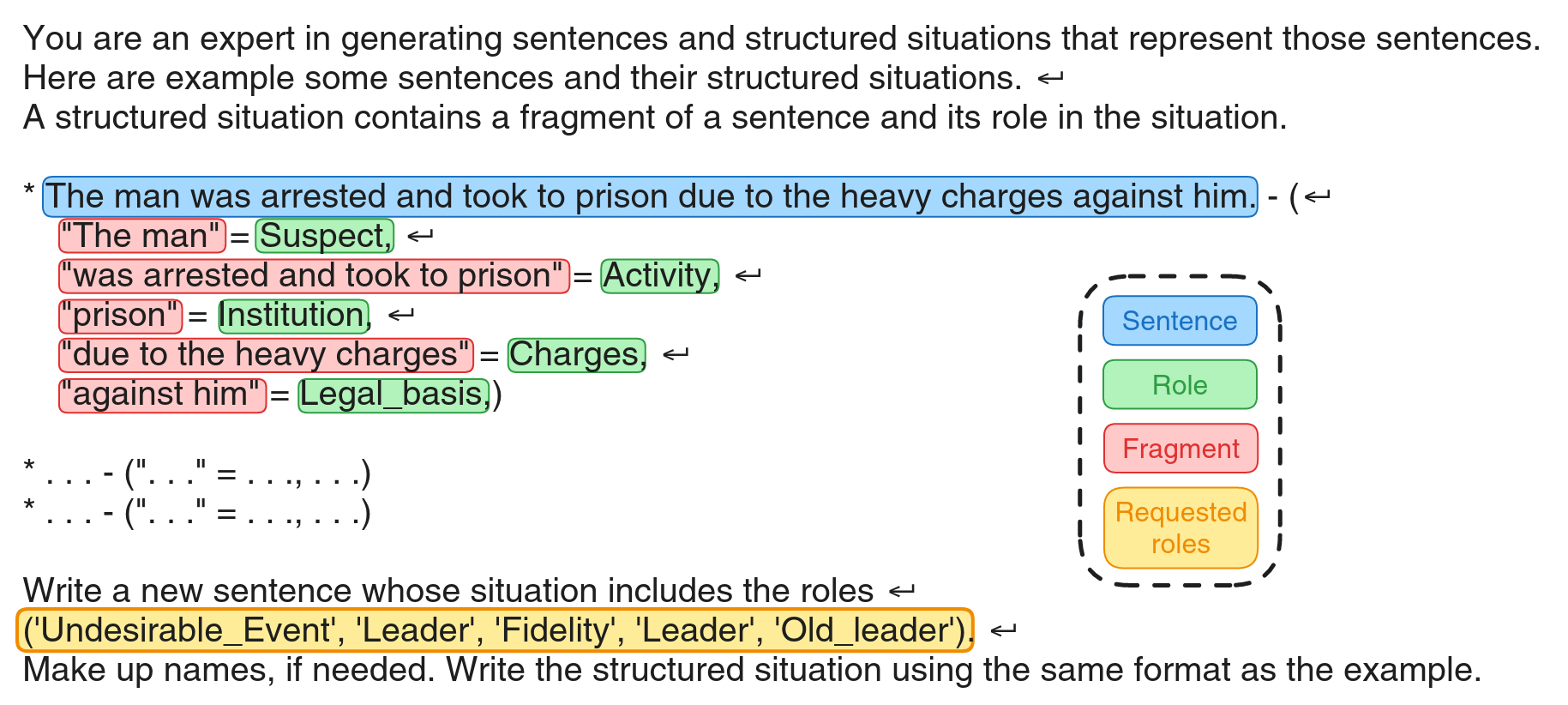}
    \caption{Annotated prompt example.}
    \label{fig:annotated-prompt}
\end{figure}

    In this section, we discuss the methodology used to generate the synthetic dataset of sentences and their structured representation, and provide some insights on its quality.
For the generation of the dataset, we rely on prompting a Large Language Model (LLM). In particular, we rely on Mistral-7B \cite{jiang2023mistral}, a 7 billion parameters open-source language model. We do make use of three main techniques: \textit{cloze}, role-playnig, and few-shot prompting \cite{liu2023prompting}. 
Few-shot prompting refers to the technique where examples of desirable outputs are provided to the LLM. They have been shown to be particularly effective in classification tasks (to obtain \textit{in context} learning).
Cloze prompting, on the other hand, refers to the technique in which the above mentioned desired output, are presented in a specific structure. In our case Figure \ref{fig:annotated-prompt} shows an example, as detailed in the followings. Finally, role-playing is a technique where
the LLM is instructed to act as a specific agent. It has been observed that this kind of technique increases the quality of the generated responses.  In our setting, we rely on it to indicate the output structure, which allows us to easily parse the sentence and its structured representation, as well as provide examples of sentences that are manually curated to express some specific value.

Figure \ref{fig:annotated-prompt} shows an example of a prompt. In the first part, we implement cloze prompting and instruct the LLM on its role as a structured sentence generator. Afterwards, we present 3 example outputs structured as a tuple composed of a sentence and its representation as a structured situation. The structured representation is composed of a fragment of the sentence and its annotated role. Finally, we ask the LLM to provide a new sentence that includes some specific roles and we instruct it to provide the results compliant to the examples we gave. The requested roles are randomly sampled in the range $[2, 15]$ from the ones in ValueNet \cite{degiorgis2022valuenet}. We extract all the roles used in every description. This allows the LLM to generate sentences whose roles are not confined to a specific value. Notice that in the prompt of Figure \ref{fig:annotated-prompt} we never explicitly mention to the LLM that the sentences to be generated have to evoke some values nor we ever mention the concept of values. This is done to avoid any bias in the generative process. The model should autonomously extrapolate the aspects that are in common between all the example sentences and generate a novel one that displays those aspects. This results in a more challenging dataset where different perspectives must be considered. It is not possible to assume that a sentence \textit{prototypically} expresses a particular value, but rather that among all the perspectives that explain it, there exists one in which the sentence can be classified with that particular value.

In total we generate $10,000$ structured sentences, $1,000$ sentence for value. The values considered are \textit{authority}, \textit{care}, \textit{cheating}, \textit{degradation}, \textit{fairness}, \textit{harm}, \textit{loyalty}, \textit{oppression}, \textit{sanctity}, \textit{subversion}. Each value is represented as a description. On average, each description is composed of $51$ roles out of a total of $513$ unique roles. 

\begin{figure}
    \centering
    \begin{subfigure}[b]{0.49\textwidth}
        \centering
        \includegraphics[width=\textwidth]{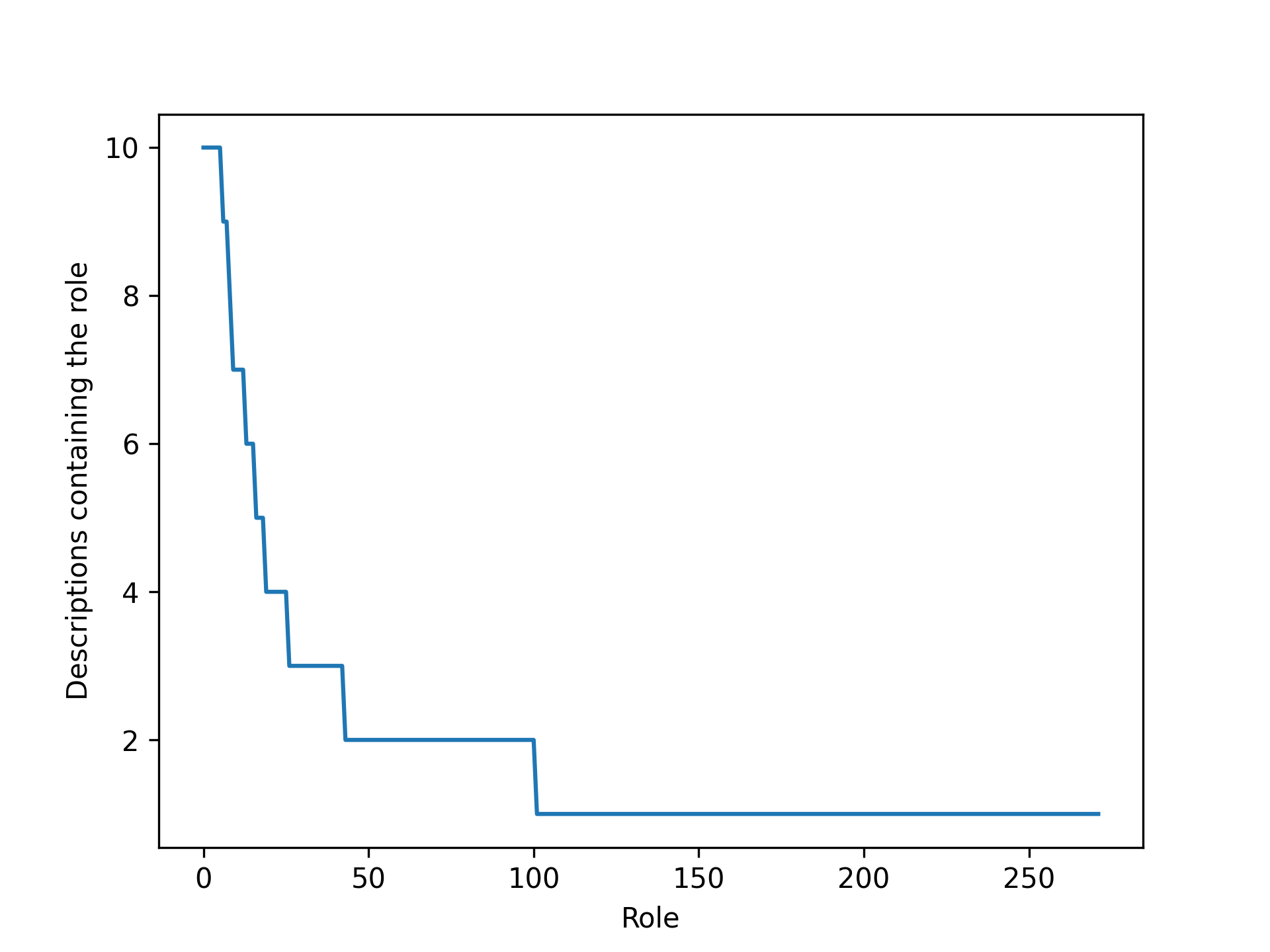}
        \caption{Distribution of the roles within ValueNet.}
        \label{fig:role-distribution-ontology}
    \end{subfigure}
    \hfill
    \begin{subfigure}[b]{0.49\textwidth}
         \centering
         \includegraphics[width=\textwidth]{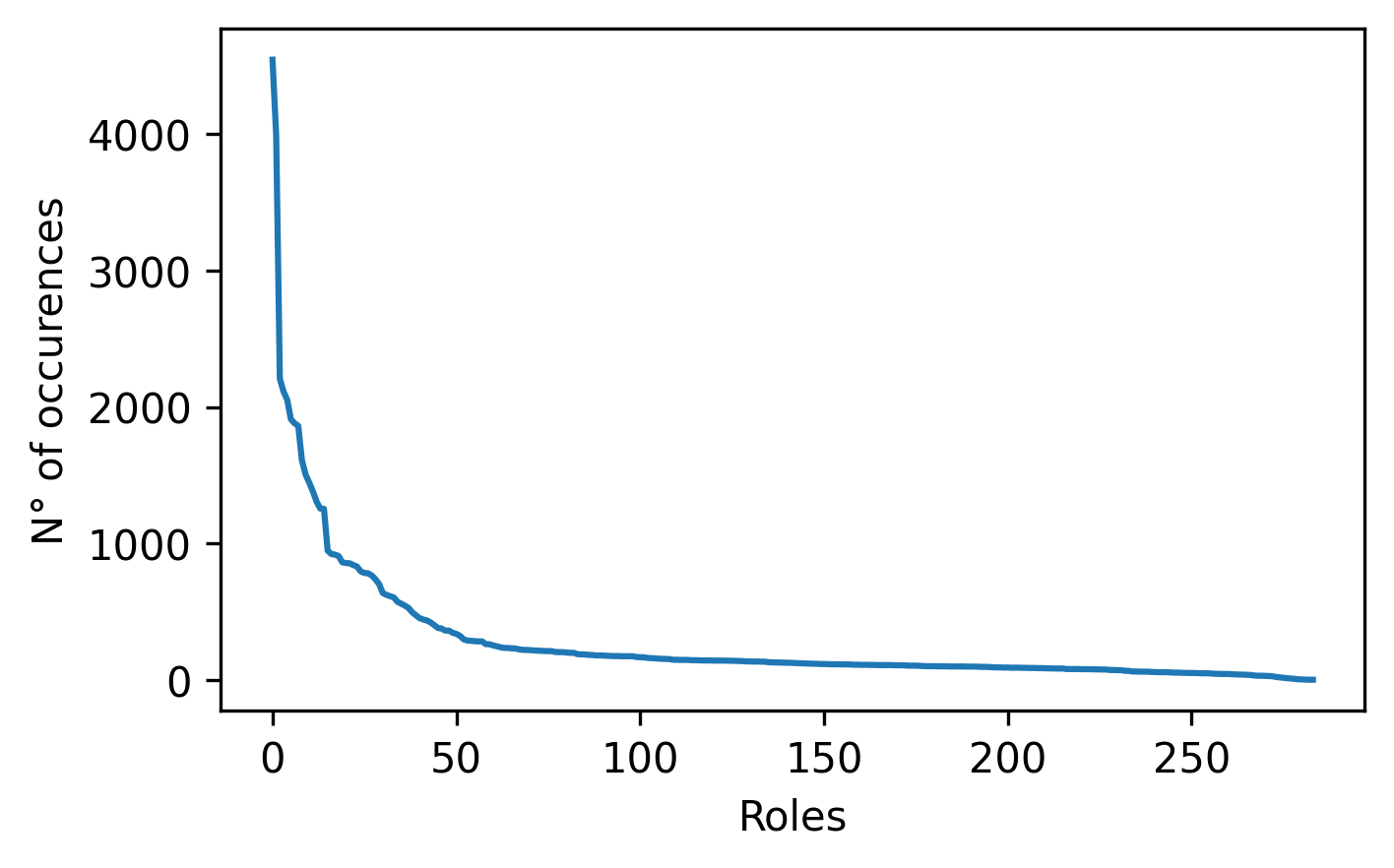}
         \caption{Distribution of the roles within the generated dataset.}
         \label{fig:role-distribution-dataset}
    \end{subfigure}
    \caption{Distribution of the roles within the ValueNet ontology and in the generated dataset.}
    \label{fig:role-distribution}
\end{figure}

The distribution of the roles within ValueNet follows the well-known Zipf distribution (see Figure \ref{fig:role-distribution}). Indeed, the most common roles (e.g. \textit{Time}) are very common concepts that characterize every possible value. To guarantee variability within the dataset, we randomly sample the \textit{temperature} parameter of the LLM in the range $t \in [0.01, 1]$.

\begin{figure}[htbp]
    \centering
    \includegraphics[width=\textwidth]{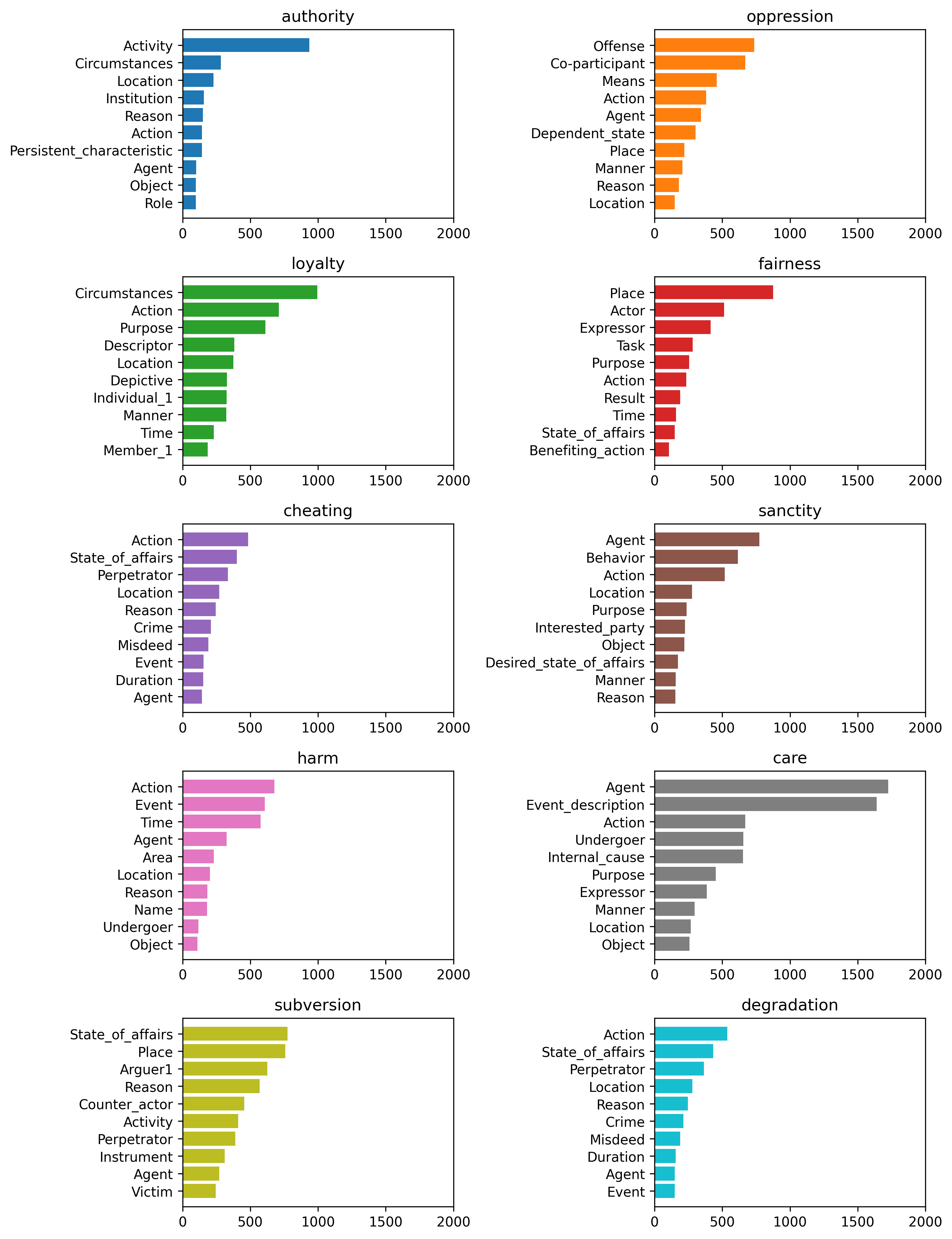}
    \caption{Top 10 roles for each value used in few shot prompting.}
    \label{fig:top10-roles-distribution}
\end{figure}

Given the prompt as in Figure \ref{fig:annotated-prompt}, the extraction of the generated sentence and of its respective situation is straightforward. Nonetheless, we note that the output produced by the LLM, despite the clear indications, is not always consistent.
From the total of $10,000$ sentence, $1,166$ sentences (more than $10\%$) contains hallucinated roles. We discard those sentence. Moreover, after some manual inspection we found that the LLM does not always faithfully comply with the proposed structure. We implement a straightforward parser, based on regex patterns, and find that only $77\%$ of the sentences have been correctly extracted (we consider a sentence correctly extracted when it is terminated by a full-stop). A more carefully implemented parser might help in reducing the amount of noise within the dataset. A total of $6844$ structured sentences is extracted after removing the ones that are wrongfully parsed. The distribution of roles within the dataset (Figure \ref{fig:role-distribution-dataset}) is similar to the one in the ValueNet (Figure \ref{fig:role-distribution-ontology}). Indeed, even though we provide a set of roles that the LLM must consider during the generation phase, it will also autonomously generate sentences that include other existing roles. The choice of those is biased towards roles that are more common within the pre-training corpus. In Figure \ref{fig:top10-roles-distribution} the top 10 roles for each value used in the few-shot template are shown. Indeed, we can see that some roles, such as \textit{Action} or \textit{Agent}, are shared among different values. Those roles represent very general concepts that are very likely to occur in a sentence, regardless of its intended meaning. On the other hand, some roles are particularly prominent on some values rather than others, such as \textit{Perpetrator} for the \textit{Cheating} value. This supports the idea that specific roles have different relevance and salience among different frames.

\begin{wrapfigure}{R}{0.5\textwidth}
    \centering
    \includegraphics[width=0.48\textwidth]{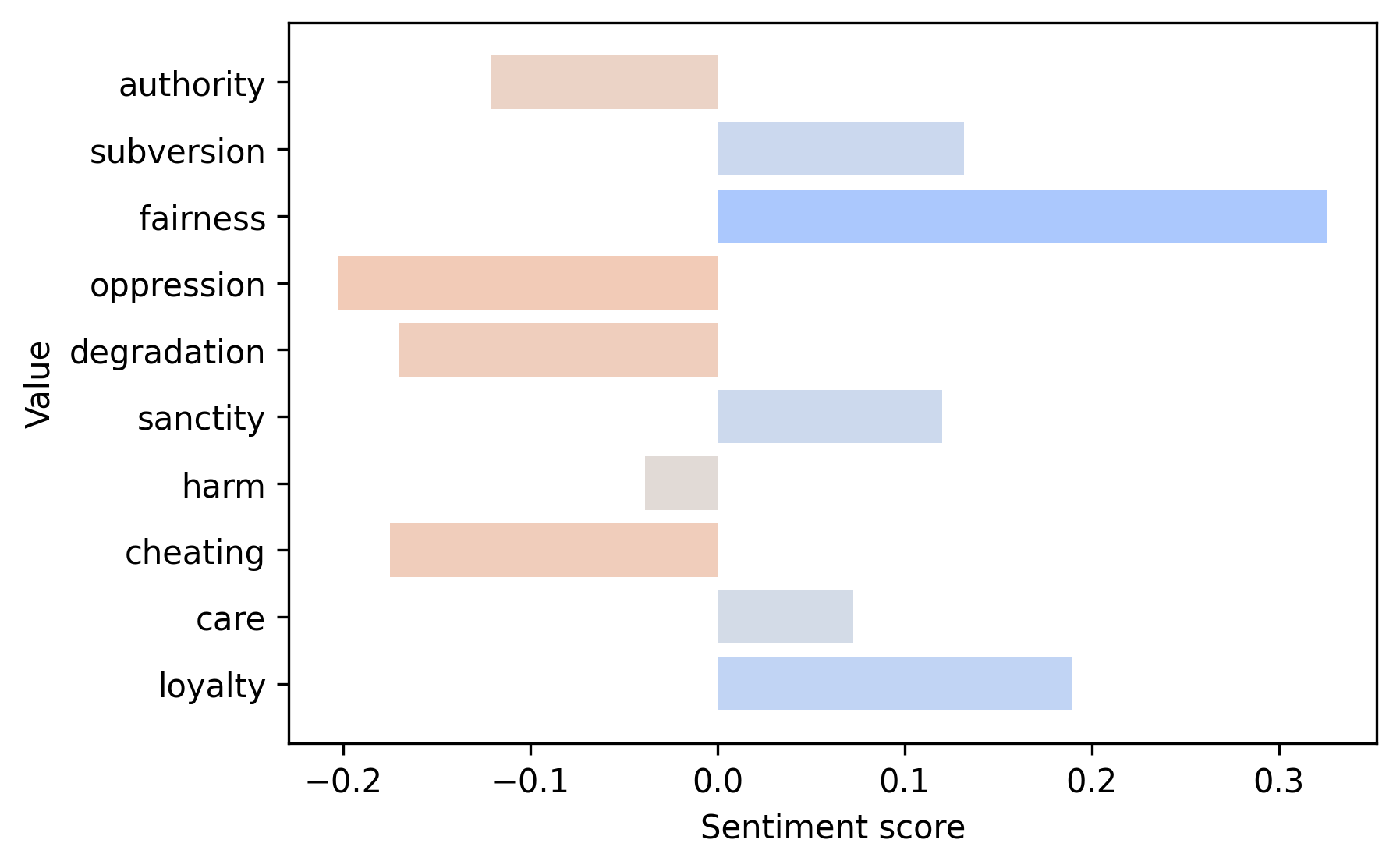}
    \caption{Average emotion polarity detected on the generated sentences for each value used in the few-shot example.}
    \label{fig:emotion-polarity}
\end{wrapfigure}

As addressed previously, it is not possible to assume that the sentence generated by the LLM when provided with few-shots for a specific values are guaranteed to evoke that specific value. Meticulous Prompt Engineering is usually involved in order to produce reliable results \cite{javaheripi2023phi,benallal2024cosmopedia}. One possible form of validation is to check whether the emotion polarity extracted from the sentences correlates with the value they are tagged as. 

Figure~\ref{fig:emotion-polarity} describes the emotion polarity, computed with NLTK \cite{bird2006nltk}, for all the sentences in the dataset grouped by the value used in the few-shot examples. It can be observed that values' polarity is coherent with those values that are \textit{prototipically} recognised as associated with ``moral emotions'', such as in the CAD (Contempt-Anger-Disgust) theory \cite{rozin_cad_1999}. E.g. values associated to positive emotions, such as \textit{fairness} and \textit{loyalty}, have a positive emotion polarity on average while values such as \textit{oppression} and \textit{cheating} have a lower polarity on average.
    
    \section{Reasoning on Descriptions and Situations}
    \label{sec:background}
    ValueNet is developed \cite{degiorgis2022valuenet} by re-using the Description and Situation (DnS) ontology design pattern (ODP) \cite{gangemi2008plansandnorms}, where knowledge is represented using a frame-based representation.
DnS introduces the concepts of \emph{description} and \emph{situation}. Both descriptions and situations are n-ary relations over a set elements. A description is a perspective (a theory, a schema) defining concepts that can classify, hence interpret, a situation. The elements that are part of a description are called \textit{roles}. In ValueNet, one description is created for each value/violation. Examples of roles include \texttt{Activity} or \texttt{Circumstances}, as can be seen in Figure~\ref{fig:top10-roles-distribution}. A situation is an n-ary relation involving a set of individuals that are classified with one role. For instance, in Figure~\ref{fig:annotated-prompt}, the sentence represented in blue is a situation in which a man is taken to prison due to some charges. The individuals that partake on that situation are its fragments, annotated in red. Each of those individuals are classified as a role (e.g. ``The Man'' is classified as the role \texttt{Suspect}).

A situation $s$ \emph{satisfies} a description $d$ when $d$ is a plausible or correct interpretation of $s$. This means that for every role $r$ in $d$, then there exists an individual in $s$ that has been classified by that role.
For example, if we assume that the \texttt{Fairness} value is formalized as a description that only involves the roles \texttt{Suspect} and \texttt{Charges}, then the sentence of Figure~\ref{fig:annotated-prompt} satisfies the description of \texttt{Fairness}.

Reasoning over DnS is performed by relying on the neuro-symbolic reasoner sandra \cite{lazzari2024sandra}. Sandra builds a geometrical space based on descriptions and roles defined in a DnS ontology and reason over it to compute the probability of each description satifying a situation.

Given the dataset produced in Section \ref{sec:data}, where each sentence's fragments are classified by one role compatible with ValueNet, is is possible to infer which descriptions are satisfied by a sentence (represented as a situation). Note that inferring which description is satisfied by a situation is a different classification problem than the value detection methods described in Section~\ref{sec:related}. In particular, if a situation satisfies a description, then it means that the situation can be interpreted through the perspective formalized by that description. Different (contrasting) perspectives can be valid at the same time.
For example, when classifying the dataset produced in Section \ref{sec:data}, we find that only $30\%$ of the sentences are predominantly satisfied by the description associated with the value used in the few-shot example. This is a consequence of the formalization in ValueNet. Different values, represented as descriptions, share many roles. As a result, it is easy to see that if a situation involves individuals that are classified by those shared roles, it will be classified by different (incompatible) descriptions with a similar probability.

\begin{figure}
    \centering
    \begin{subfigure}[b]{0.49\textwidth}
        \centering
        \includegraphics[width=0.7\textwidth]{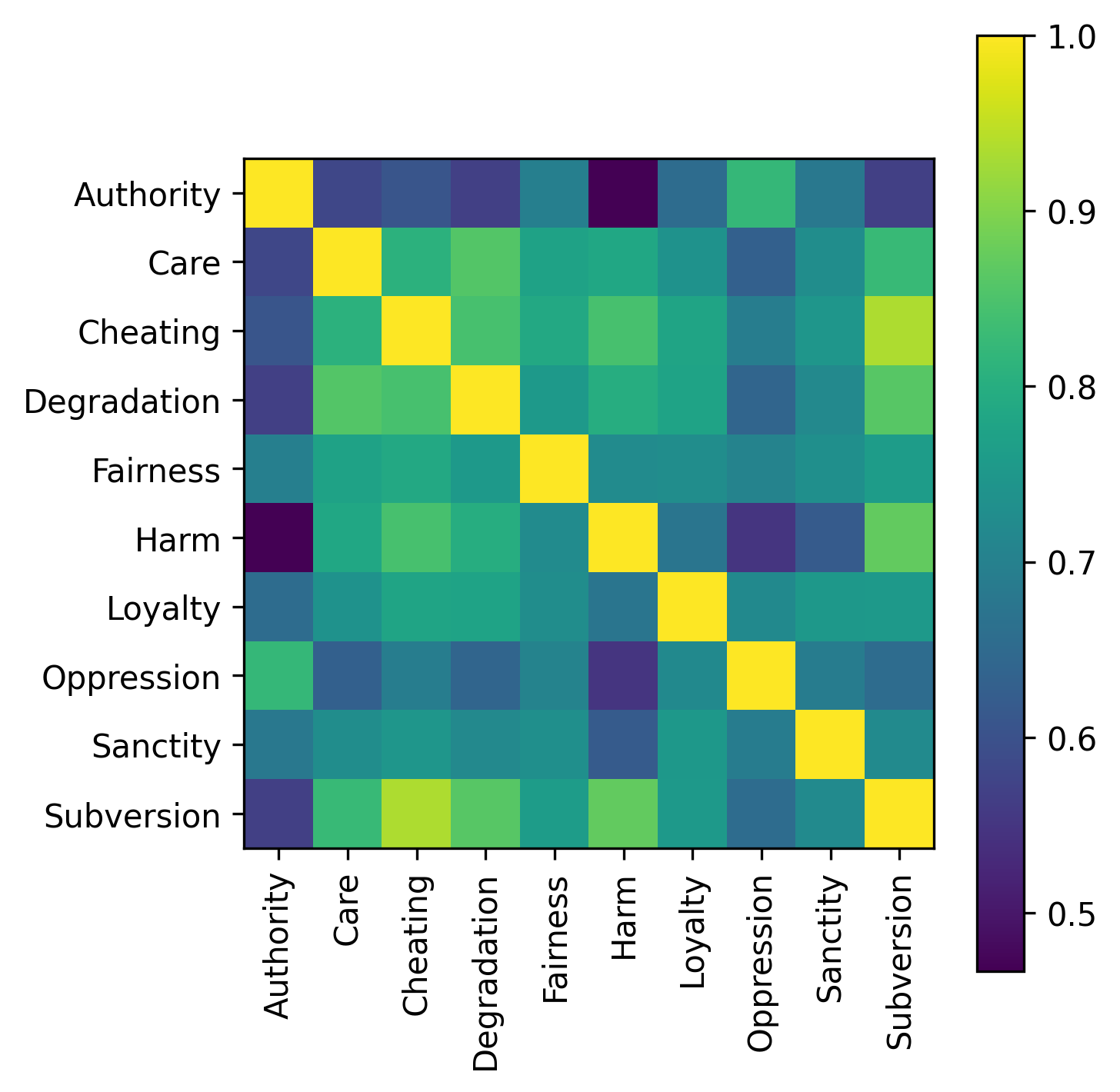}
        \caption{Pearson correlation between the satisfaction of different description by a situation.}
        \label{fig:description-sat-corr}
    \end{subfigure}
    \hfill
    \begin{subfigure}[b]{0.49\textwidth}
         \centering
         \includegraphics[width=0.8\textwidth]{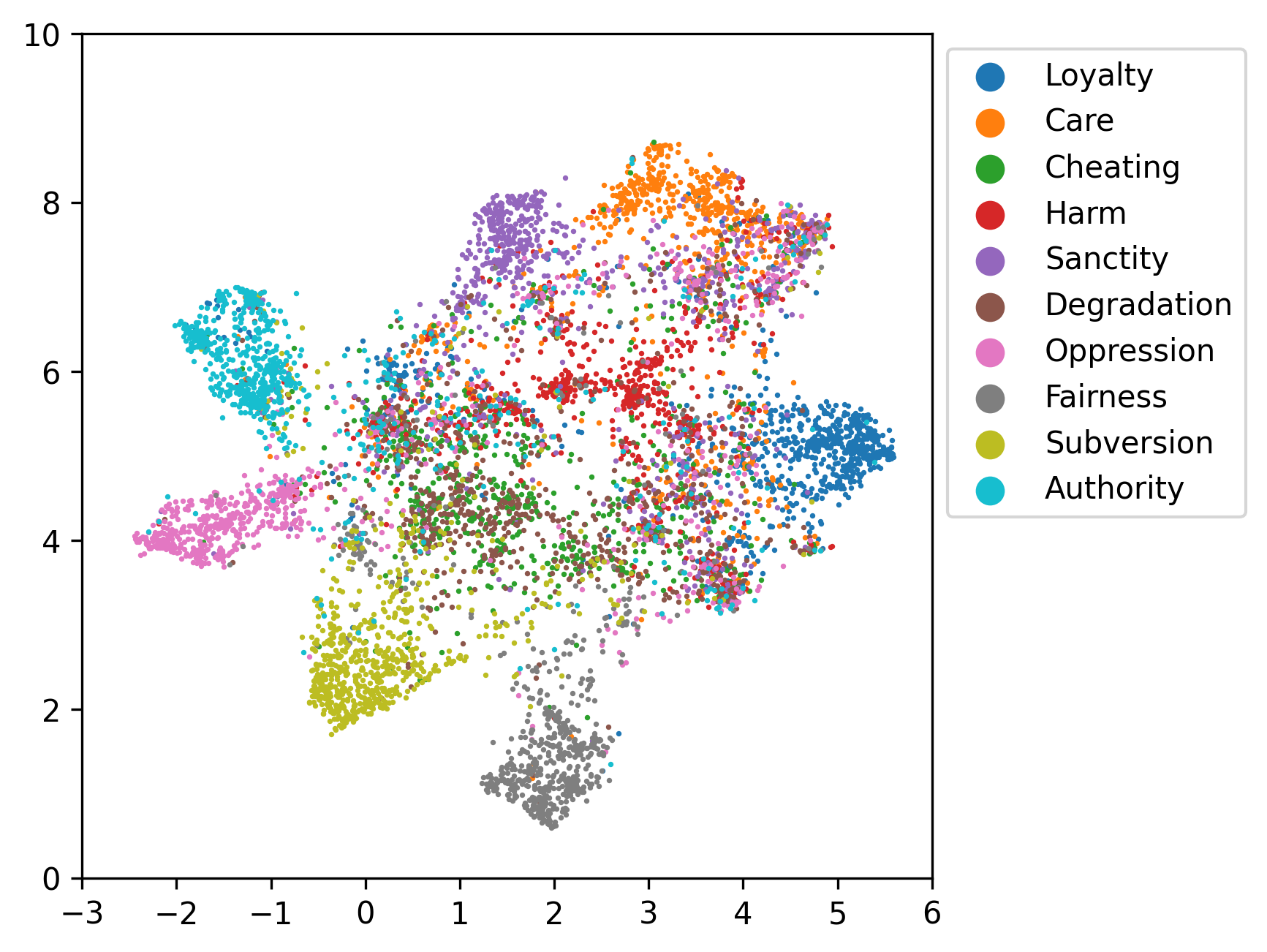}
         \caption{Two dimensional plot of the situations where the probabilities of satisfying each description is used to compute the position in the plane.}
         \label{fig:2d-plot-satisfaction}
    \end{subfigure}
    \caption{Analysis of the description satisfied by each situation.}
    \label{fig:satisfaction-analysis}
\end{figure}

Figure~\ref{fig:satisfaction-analysis} provides additional insights that can be obtained by classifying the situations using descriptions.
In particular, in Figure~\ref{fig:description-sat-corr} the Pearson correlation between the different descriptions is shown. A description is highly correlated with another description if, when considering multiple situations, it is often the case that both descriptions have \textit{similar} probabilities. It can be seen that the positive correlations (an increase in probability on one description results in an increase on the other as well) between some descriptions has an intuitive explanation. For instance, \texttt{Sanctity} correlates with \texttt{Loyalty}, but not with \texttt{Harm}. On the other hand, some cases are surprisingly interesting. For example, \texttt{Care} is correlated with \texttt{Degradation} and \texttt{Subversion}. From a technical point of view this might be because the roles that can be found in \texttt{Care} are also in common with \texttt{Degradation} and \texttt{Subversion}. 
From a conceptual point of view, it could point in the direction that \texttt{Degradation} and \texttt{Subversion} could be often expressed via a negation or lack of \texttt{Care}.
Another interesting case is \texttt{Oppression}, highly correlating with \texttt{Authority}, coherently with the idea that restrictions of individual freedom are usually felt as a general sense of oppression, when backed by some form of institution or systemic provenance.
These unexpected correlations provide an indication that either ValueNet lacks of granularity, or rather that the formalization of values such as \texttt{Degradation} and \texttt{Subversion} could be implicitly related to presence or absence of other values in the sentence.
In Figure~\ref{fig:description-sat-corr}, the probabilities of satisfying a description is used as a feature vector and plotted in two dimensions through the UMAP dimensionality reduction method \cite{sainburg2021umap}. Notably, situations that are generated using examples from a specific values are closely positioned in the low-dimensional space for most of the values. Some values, such as \texttt{Harm}, \texttt{Cheating} and \texttt{Degradation}, are not neatly separated. This further support the arguments described for Figure~\ref{fig:description-sat-corr} related to ValueNet's definitions.
    
    \section{Value Detection Experiments and Discussion}
    \label{sec:method}

As anticipated in Section \ref{sec:background}, the ability to infer which descriptions are satisfied by a situation should not be framed as the same classification task that traditional value detectors perform (c.f. Section \ref{sec:related}). The ability to infer which descriptions are valid perspectives for a situation shall be interpreted as an information upon which the final classification must be performed. In this section, we obtain a feature vector by concatenating the probability that a given situation is satisfied by every description in ValueNet.

There are two possible ways of exploiting this feature vector:
\begin{itemize}
    \item classify a sentence \underline{only} based on the descriptions it satisfies;
    \item rely on the feature vector to \underline{complement} other features extracted from the text.
\end{itemize}

We asses both methods by training several popular ML classification methods to classify the value of a sentence. To do so, we randomly split the sentences in training and testing sets, keeping two thirds of the situations to train the models and one third to test it. To assess the influence of using sandra as complementing features, we follow two different feature extraction methods from text: TF-IDF and sentence-embeddings (sBERT) \cite{reimers2019sentencebert}\footnote{We use the model available at \url{https://huggingface.co/sentence-transformers/all-mpnet-base-v2}}.

\begin{table}[ht]
    \centering
    \caption{Accuracy results obtained with different methods and different feature extraction algorithms. Best results for each method are underlined while the best result overall is represented in bold.}
    \label{tab:results}
    \begin{tabular}{lccccccc}
        \toprule
        Model & sandra & TF-IDF & TF-IDF + sandra & sBERT & sBERT + sandra \\ \midrule
        Logistic regression & 0.38 & 0.49 & \underline{0.53} & 0.45 & 0.49 \\
        SVM & 0.48 & 0.48 & \underline{0.56} & 0.45 & 0.52 \\
        Random forest & 0.48 & 0.44 & \underline{0.57} & 0.34 & 0.45 \\
        Decision Tree & 0.39 & 0.29 & \underline{0.42} & 0.20 & 0.32 \\
        LightGBM & 0.50 & 0.45 & \underline{\textbf{0.62}} & 0.42 & 0.57 \\
        MLP & 0.49 & 0.45 & 0.47 & 0.44 & \underline{0.52} \\
        Naive Bayes & 0.29 & \underline{0.43} & 0.42 & 0.32 & 0.32 \\
        KNN & \underline{0.45} & 0.33 & 0.33 & 0.32 & 0.32 \\
        \bottomrule
    \end{tabular}
\end{table}

Table \ref{tab:results} describes the results obtained on the dataset presented in Section \ref{sec:data}. Notably, those relying on sandra outperforms all the other methods. In particular, complementing a word-based representation, TF-IDF, with inference supported by a formal representation yields the best results with 5 out of 8 completely different methods. Interestingly sBERT, which relies on dense representation that benefits from a large pre-training phase, does almost never outperform the other models with the exeception of when being used in conjuction with sandra and an MLP classification layer. This further provides evidences that complex tasks that require a higher semantic understanding of a sentence are better solved by integrating representations that rely on the distribution of data (TF-IDF and sBERT) with representations that rely on a formal theory.
We remark that out of the model used in Table \ref{tab:results}, some are explainable models, i.e. a human-understandable explanation can be obtained from the decision process used to classify a sentence.
In the following, we will provide some examples of the type of explanations that can be obtained from some of the models presented.

\subsection{Logistic regression}
The logistic regression classification method is a linear model that seeks to classify a feature vector by computing a weighted sum of the features and using a logistic function (i.e. a function that squashes values between 0 and 1) to produce a probability over each class. Hence, given the set of values we are interested in, the model learns the right weight for each feature in order to produce a correct probability score. The weights can be manually inspected to understand which features are the most important to predict one label instead of another.

\begin{figure}[htb]
    \centering
    \includegraphics[width=1\linewidth]{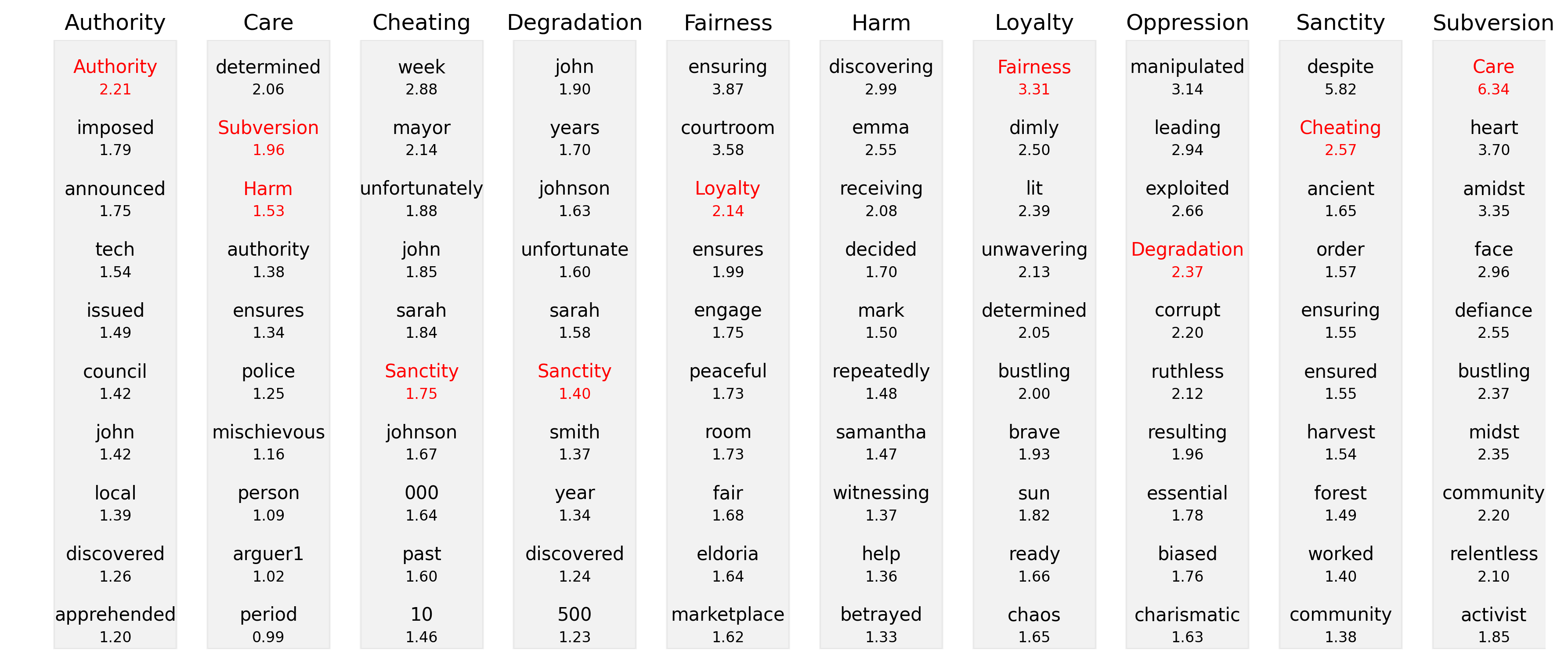}
    \caption{Feature importance for each value in the dataset. For each value, the top 10 features by weight are reported. The descriptions from ValueNet and their score are represented in red.}
    \label{fig:feature-importance-regression}
\end{figure}

In Figure \ref{fig:feature-importance-regression} the importance of each feature for the logistic regression model using TF-IDF and sandra is reported. It can be seen how, as also discussed previously, the fact that a situation satisfies a description directly modeled after a value does not entail a classification result. Indeed, only for the \texttt{Authority} value the respective description is among the top 10 most important features. In general, we can see that words are not always informative. For example, the word \texttt{john} is an important feature for many values. However, this is not correlated with a value, but is rather a result of the generation method used in Section \ref{sec:data}. Excluding words that should not be taken into consideration for this task (e.g. proper nouns) and only mantaining verbs and concepts could greatly enhance the explainability of the model and its performances. Interestingly, for values in which satisfying another value is important in the decision process (e.g. \texttt{Loyalty}), the emotion polarity of both values is the same, in this case \texttt{Loyalty} and \texttt{Fairness} are both generally considered positive values. On the other hand it can be seen that for \texttt{Care} the \texttt{Subversion} and \texttt{Harm} descriptions are particularly important. There might be multiple reasons that explain this results, one of which includes the fact that there exist a possible perspective from which a situation satisfying a \texttt{Harm} description might actually satisfy a \texttt{NotCare} description, where \texttt{NotCare} is a possible perspective from which \textit{harm} as a value can be interpreted. This further supports the claim that the ValueNet ontology might benefit from further refinement in terms of granularity. For example, identifying possibly conflicting perspectives from which the same value can be seen.

\subsection{LightGBM}

\begin{wrapfigure}{hR}{0.5\textwidth}
    \centering
    \includegraphics[width=0.45\textwidth]{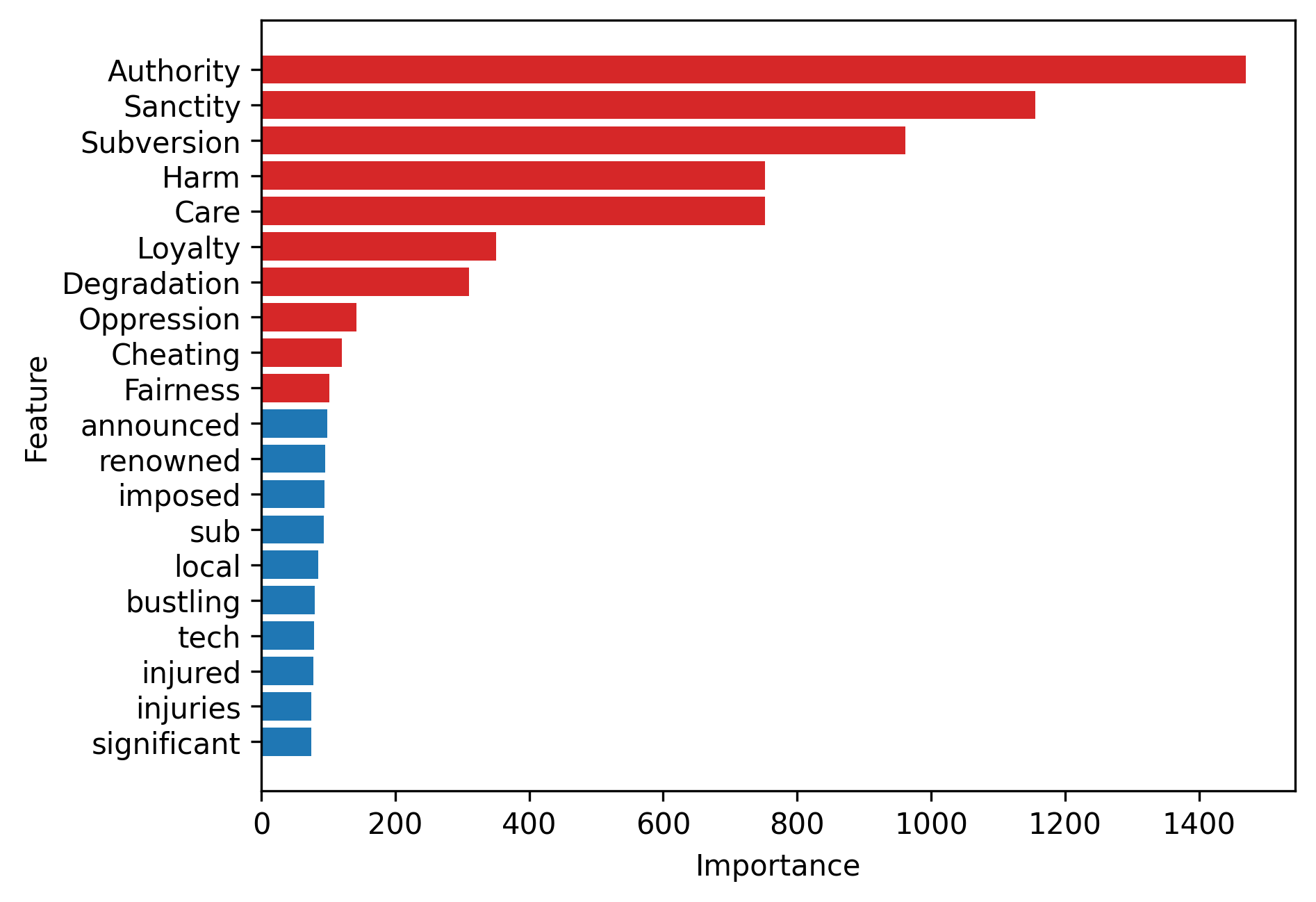}
    \caption{Importance of the top 20 features in the LightGBM model.}
    \label{fig:feature_importance_gbm}
\end{wrapfigure}

The LightGBM method \cite{ke2017lightgbm}, a gradient boosting classification technique, is currently the state-of-the-art method in classification of tabular data. It is widely used in industrial Machine Learning applications, due to its robustness and accuracy in achieving good results. Being a gradient boosting method, LightGBM is based on the use of multiple decision trees. A decision tree is a method that construct a binary tree based on the input feature. By taking into account one feature at the time, decision trees seek to find the threshold such that a feature can be used to make a binary decision. For instance, a binary decision might be to check for the presence of a word given its TF-IDF score, or checking if the probability of satisfying a description is higher (or lower) than a specified threshold. During this procedure, the most important features are identified. These are the features that mostly help in reducing the set of choices for the final classification.

In Figure \ref{fig:feature_importance_gbm} the importance of each feature is plotted. It can be seen that, similarly for the logistic regression model, the importance of each situation satisfying (or not) a description is fundamental to make an accurate prediction. It can also be seen that some descriptions are much more important than others, for instance \texttt{Authority} is one order of magnitude more important than \texttt{Oppression} or \texttt{Fairness}. This is still to be reconducted to the ValueNet formalization, as the same value as been proven to be very important also in the logistic regression model.
Note that in the case of LightGBM, it is possible to obtain accurate performances through the use of multiple (usually in the order of tens of thousands) decision trees. However, it is not possible to directly understand the decision process of each tree, as it would be hard to evaluate each of them single-handedly. From Table \ref{tab:results}, it can be seen that a single Decision Tree model does not perform well compared to the gradient boosting technique. Nonetheless, by focusing on improving the ontology to optimize the classification performances while retaining a well-formalized theory, it might be possible to obtain a fully-explainable model which is easy to understand by humans and obtains accurate performances, comparable to more complex methods.

    \section{Implementation}
    \label{sec:implementation}
    \begin{figure}[htbp]
    \centering
    \begin{subfigure}[b]{0.45\textwidth}
        \includegraphics[width=.9\textwidth]{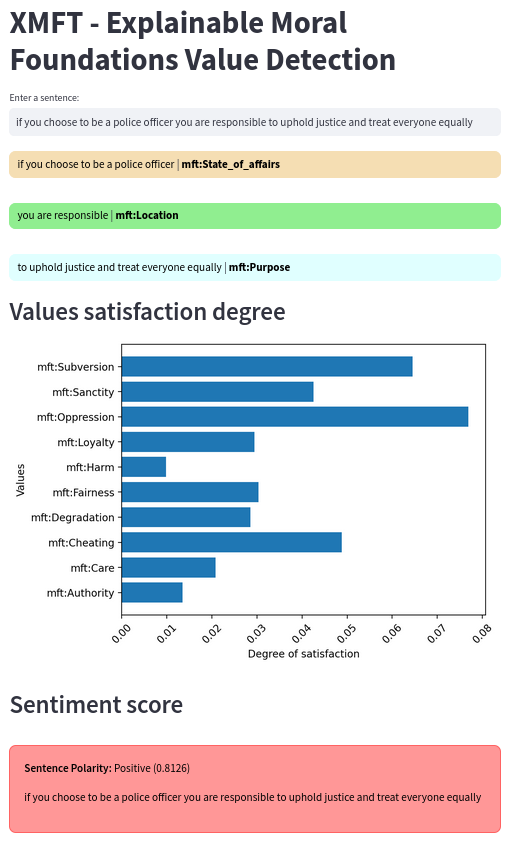}
        \caption{Visualization of the sentence \textit{If you choose to be a police officer your are responsible to uphold justice and treat everyone equally}.}
    \end{subfigure}
    \begin{subfigure}[b]{0.45\textwidth}
        \includegraphics[width=.9\textwidth]{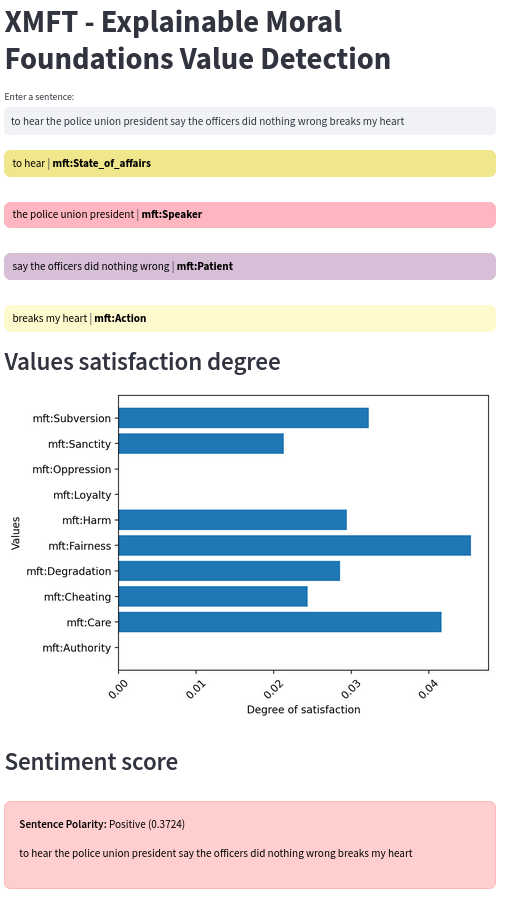}
        \caption{Visualization of the sentence \textit{To hear the police union president say the officers did nothing wrong breaks my heart}.}
    \end{subfigure}
    \caption{Example of visualizations from sentences extrapolated from the Moral Foundation Reddit Corpus \cite{trager2022mftcorpus}.}
    \label{fig:viz}
\end{figure}

In order to provide a comprehensive and accessible demonstration of our approach, we have developed an online service that visualizes and generalizes our methodology, shown in Figure \ref{fig:viz}. This web-based platform is designed to make the detection of moral values from natural language both transparent and interactive. The motivation behind this implementation is rooted in the need for explainable AI systems, especially in the context of moral value detection, which is inherently subjective and complex. By providing a visual and interactive tool, we aim to bridge the gap between theoretical models and practical applications, ensuring that users can clearly understand and trust the results generated by our system.

Our approach leverages the Moral Foundations Theory (MFT) to detect values in text. The ability to automatically discern MFT values from text is significant for understanding human behavior, decision-making processes, and societal trends. However, the challenge lies in making these detections interpretable and explainable. Our online service addresses this challenge by providing a clear visualization of the data and the methodology described in previous sections.

\subsection{Technical Framework}
\label{sec:technical_framework}

The interface of our online service, available at \url{http://xmv.geomeaning.com}, as depicted in Figure \ref{fig:viz}, is designed with usability and clarity in mind, as well as explainability of the value detection process. Users are prompted to input a string in natural language, which initiates the processing sequence. At the core of our system is a BART-based semantic role labeling model. BART \cite{lewis2020bart} is a model known for its high performance in generating contextually accurate and coherent labels on machine translation tasks. We chose BART due to its effectiveness in handling complex NLP tasks and its ability to produce reliable role annotations critical for our value detection process.

\subsubsection{BART Role Labeling}
\label{sec:bart}

The role labeling process involves assigning specific roles to segments of the input text, which are then used to determine the satisfaction of various moral values. The roles are taken directly from the MFT ontological module of the ValueNet ontology, as described  in Section \ref{sec:method}, ensuring that the detected values are grounded in a well-established theoretical framework. This step is crucial as it transforms the input text into a structured format that can be further analyzed.
Possible downsides of this approach are described ins Section \ref{sec:final_remarks}.

\subsubsection{Value Descriptions Satisfaction}
\label{sec:value_satisfaction}

Once the roles are labeled, the system generates histograms that visualize the satisfaction of different moral values. These histograms represent how well the input text aligns with the descriptions of various moral values. Note that a sentence that highly satisfies a moral value is not to be considered as classified by that value, but rather that its semantic representation can be partially interpreted through the formal definition of that value. The visual representation includes bars indicating the degree of satisfaction for each value, providing an intuitive and easily interpretable overview. This helps users quickly grasp which moral values are prominent in the text and to what extent.

\subsubsection{Sentiment Analysis Visualisation}
\label{sec:sentiment_viz}

This step adds a layer of meaningful correlation with emotions. We have integrated a sentiment analysis component using the sentiment.Vader package. This step analyzes the emotional tone of the text and visualizes the results on a color scale ranging from red (positive sentiment) to blue (negative sentiment). By combining sentiment analysis with moral value detection, users gain a glimpse of an interconnected view of the text's moral and emotional landscape. This dual analysis enhances the interpretability of the results, showing not only which values are present but also the sentiment associated with them.

\subsection{Final Remarks}
\label{sec:final_remarks}


The primary advantage of our implementation is its focus on explainability. By visualizing the data and providing interactive features, users can explore the results in depth. The system allows users to understand how different parts of the text contribute to the overall moral value detection, making the process transparent and understandable.
One notable consequence of this approach, is that the ``Non-moral'' label, often included in value-annotated datasets such as MFRC \cite{trager2022mftcorpus} is by design not included in this approach, due to the structure of \emph{quasi-satisfaction} based on presence or absence of certain general roles such as \emph{Agent}, \emph{Patient}, etc.
Future enhancements include the integration of the classification pipeline, devised in Section \ref{sec:method}, within the visualization dashboard. Additionally, incorporating user feedback mechanisms could help in continuously improving the system's accuracy and usability. One possible application of these kind of visualization systems is in the annotation process performed by human subject. Since the tool automatically proposes an interpretation of the sentence by identifying semantic roles within a sentence, it is possible for the user to refine the element identified, effectively relying on the system as a support to the semantic annotation of sentences. Moreover, by exploiting the inferences performed by sandra, it is possible to understand how the semantic annotation influences the possible perspectives that can be applied to that sentence. The user can hence classify the sentence by taking into account those perspectives, or improve over the semantic annotations to obtain the desired inferences. This approach paves a new way of data curation and annotation for moral values. The cognitive load over the user is reduced thanks to the support of the AI model. This allows the annotator to focus on the core of the annotation task.



    \section{Conclusion}
    \label{sec:conclusion}

In this work we reused an ontological formalization of the domain of values, in particular the Moral Foundations Theory, modeled as value frames in ValueNet. We produced a synthetic dataset of 10.000 sentences, 1k for each value considered, and reused \emph{sandra}'s reasoning approach to extract semantic roles salience and roles overlaps for each considered value. We furthermore performed preliminary value detection experiments with classic ML methods to classify the degree of satisfaction for each sentence per each value, demonstrating an improvement in performance when adopting \emph{sandra}'s method. We show that relying on a neuro-symbolic approach allows the classification process to be explainable and understandable, since the injection of satisfied description greatly improves the performances of interpretable and explainable models. Moreover, we found that by relying on ValueNet for the classification process, it is possible to identify its gaps and limitations that can be operatively addressed by an expert to produce a better formalization. Nonetheless, by relying on a synthetic dataset, our approach suffers from the bias of the LLM used to generate sentences. As described in the previous section, it is not possible to assume that the dataset is representative of real-world sentences expressing human values. Future works include, on one side, testing our methods on well established datasets, that have been validated by humans. Similarly, the proposed approach can be used to inform the refinement of the knowledge in ValueNet, particularly in enriching semantic relations among values, to further improve performance and explainability with \emph{sandra}'s approach.
    
\section*{Acknowledgments}
This project has received funding from the FAIR – Future Artificial Intelligence Research foundation as part of the grant agreement MUR n. 341, code PE00000013 CUP 53C22003630006.


\bibliography{bibliography}


\end{document}